\documentclass{jlcl}

\newcommand{\authornames}{Christian Reul, Uwe Springmann, Christoph Wick, Frank Puppe}

\newcommand{\authorlastnames}{Reul, Springmann, Wick, Puppe}

\newcommand{\articletitle}{Improving OCR Accuracy on Early Printed Books by combining Pretraining, Voting, and Active Learning}

\newcommand{\shortarticletitle}{Combination of Pretraining, Voting, and Active Learning for OCR}

\usepackage[utf8]{inputenc}
\usepackage{todonotes}
\usepackage{url}
\usepackage{multirow}
\usepackage{hyperref}
\usepackage{amsmath}
\hypersetup{colorlinks,allcolors=.,urlcolor=blue}

\newcommand{\jlclvolume}{xx}

\newcommand{\jlclnumber}{x}

\newcommand{\jlclyear}{2018}

\title{\articletitle}
\author{Christian Reul\\
Universität Würzburg\\
\texttt{christian.reul@uni-wuerzburg.de} \and
Uwe Springmann\\
digital consulting\\
 \texttt{uwe@springmann.net}}
 
\begin{document}

\setcounter{page}{1}
\thispagestyle{firstpage}

\authordata


\section*{Abstract}
We combine three methods which significantly improve the OCR accuracy of OCR models trained on early printed books:
(1) The pretraining method utilizes the information stored in already existing
models trained on a variety of typesets (\textit{mixed models}) instead of starting the training from scratch. (2) Performing cross fold training on a single set of ground truth data (line images and their transcriptions) with a single OCR engine (OCRopus) produces a committee whose members then vote for the best outcome by also taking the top-N alternatives and their intrinsic confidence values into account.
(3) Following the principle of maximal disagreement we select additional training lines which the voters disagree most on, expecting them to offer the highest information gain for a subsequent training (active learning). 
Evaluations on six early printed books yielded the following results: 
On average the combination of pretraining and voting improved the character accuracy by 46\% when training five folds starting from the same mixed model. This number rose to 53\% when using different models for pretraining, underlining the importance of diverse voters. Incorporating active learning improved the obtained results by another 16\% on average  (evaluated on three of the six books). Overall, the proposed methods 
lead to an average error rate of 2.5\% when training on only 60 lines. Using a substantial ground truth pool of 1,000 lines brought the error rate down even further to less than 1\% on average.

\section{Introduction}
Recent progress on OCR methods using recurrent neural networks with LSTM architecture \citep{hochreiter1997long} enabled effective training of recognition models for both modern (20th century and later) and historical (19th century and earlier) manuscripts and printings \citep{fischer2009automatic, breuel, springmann2014, springmannluedeling2017}. Individually trained models regularly reached character recognition rates of about 98\% for even the earliest printed books. The need to train individual models in order to reach this level of recognition accuracy for early printings sets the field of historical OCR apart from readily available (commercial and open-source) \textit{general} (also called \textit{polyfont} or \textit{omnifont}) models trained on thousands of modern fonts which yield better than 98\% recognition rates on 19th century printings and better than 99\% on modern documents. Training historical recognition models on a variety of typesets results in \textit{mixed models} which may be seen as a first approximation to modern polyfont models, but their predictive power is considerably lower than that of individual models. 

In view of the mass of available scans of historical printings we clearly need automatic methods of OCR which in turn require good historical polyfont models. As long as these models are not available and at present cannot be easily constructed (we lack the necessary historical fonts to be able to synthesize large amounts of training material automatically), our next best approach is to maximize the recognition rate of a small amount of manually prepared ground truth (GT). This is the subject of the present paper which applies the methods of pretraining, voting, and active learning (AL) to the field of historical OCR. Using an already trained model as a starting point for subsequent training with additional individual material requires the capability to add specific characters not previously included in the symbol set (the \textit{codec}) and the dynamic expansion (and reduction) of the output layer of the neural network. In the context of recurrent neural networks this was recently made possible by Christoph Wick\footnote{\url{https://github.com/ChWick/ocropy/tree/codec_resize}} as reported in \cite{reul2017transfer}. Voting is a well known method of classifier combination resulting in less errors than the best single classifier output \citep{rice1992report}. Active learning ensures that lines showing maximal disagreement among classifiers are included in the training set to enable the maximal learning effect. While more training data is always better, combining these three methods results in a level of recognition accuracy that could otherwise only be reached by a much larger amount of GT and therefore a much larger manual effort to generate it.

Chapter~\ref{section:relwork} summarizes the extensive corpus of related work for each of the three methods. In Chapter~\ref{section:matmet} we describe the printing material which the experiments of Chapter~\ref{section:exp} are based on. Chapter~\ref{section:disc} contains the discussion of our results and we conclude the paper with Chapter~\ref{section:con}.

\section{Related Work}
\label{section:relwork}
In this section we first sum up a selection of important contributions concerning OCR relevant to our task and introduce findings with regard to training and applying mixed models using OCRopus. Next, a brief summary of the history of voting techniques and applications in the field of OCR is provided. After a short section on transfer learning we give an overview over some basic AL concepts.

\subsection{OCR and Mixed Models}
\cite{breuel} used their own open source tool OCRopus\footnote{\url{https://github.com/tmbdev/ocropy}} to recognize modern English text and German Fraktur from the 19th century by training mixed models, i.e. models trained on a variety of fonts, typesets, and interword distances from different books. The English model was trained on the UW-III data set\footnote{\url{http://isis-data.science.uva.nl/events/dlia/datasets/uwash3.html}} consisting of modern English prints. Applying the model to previously unseen lines from the same dataset yielded a character error rate (CER) of 0.6\%. The training set for the Fraktur model consisted of mostly synthetically generated and artificially degraded text lines. The resulting model was evaluated on two books of different scan qualities yielding CERs of 0.15\% and 1.37\%, respectively. 

\cite{ul2013can} promoted an approach not only mixing different types but also various languages by generating synthetic data for English, German, and French. Apart from three language specific models they also trained a mixed one. While the language specific models obviously performed best when applied to test data of the same language yielding CERs of 0.5\% (English), 0.85\% (German) and 1.1\% (French) the mixed model also achieved a very low CER of 1.1\% on a mixed dataset. These experiments indicate a certain robustness or even independence of the OCRopus LSTM architecture regarding different languages in mixed models.

After proving that OCR on even the earliest printed books is not only possible but can be very precise \citep[down to 1\% error rate,][]{springmann2015tutorial}, Springmann et al. adapted the idea of training mixed models to early prints in different application scenarios. In \cite{springmann2016automatic} their corpus consisted of twelve Latin books printed with Antiqua types between 1471 and 1686. Training on one half of the books and evaluating on the other half mostly yielded CERs of under 10\%. Obviously, these results are far off the numbers reported above which can be explained due to the vastly increased variety of the types. Still, the trained models provide a valid starting point for further model improvements through individual training. Additionally, a clear correlation between the intrinsic confidence values of OCRopus and the resulting CER was demonstrated.

In \cite{springmannluedeling2017} a similar experiment was conducted on the 20 German books of the RIDGES Fraktur corpus\footnote{\url{http://korpling.org/ridges}}. Again, by training mixed models on half of the books and evaluating on the held-out data impressive recognition results of around 5\% CER on average were achieved. As expected, the individually trained models performed even better, reaching an average CER of around 2\%.

\subsection{Alignment and Voting}
\label{sec:RW_vote}
\cite{handley1998improving} gives an overview regarding topics concerning the improvement of OCR accuracy through the combination of classifier results and discusses different methods to combine classifiers and string alignment approaches.

\cite{rice1996isri} released a collection of command line scripts for the evaluation of OCR results called the ISRI Analytic Tools. Their voting procedure first aligns several outputs using the Longest Common Substring (LCS) algorithm \citep{rice1994algorithm} and then performs a majority vote. In several competitions they applied their tools to evaluate the results of various commercial OCR engines on modern prints \citep[see, e.g.,][]{rice1992report, rice1996fifth}. By voting on the output of five engines on English business letters the character accuracy rate ($\text{CAR} = 1 - \text{CER}$) increased from between 90.10\% and 98.83\% to 99.15\%.

A simple but effective way to achieve variance between the voting inputs was proposed by \cite{lopresti1997using} by simply scanning each page three times. While using only a single OCR engine they still achieved a reduction of error rates between 20\% and 50\% on modern prints resulting in a CAR of up to 99.8\%. 

\cite{boschetti2009improving} improved the output of the best single engine (ABBYY, up to 97\% CAR) by an absolute value of 2.59 percentage points by applying a Naive Bayes classifier on the aligned output of three different engines on Ancient Greek editions from the 19th and 20th century. Beforehand, they performed a progressive alignment which starts with the two most similar sequences and extends the alignment by adding additional sequences.

\cite{lund2011progressive} used voting and dictionary features as well as maximum entropy models trained on synthetic data. Applied to a collection of typewritten documents from Word War II they recorded a relative gain of 24.6\% over the word error rate of the best of the five employed OCR engines.

An approach for aligning and combining different OCR outputs applicable to entire books was introduced by \cite{wemhoener2013creating}. First, a pivot is chosen among the outputs. Then, all other outputs are aligned pairwise with the pivot by first finding unique matching words in the text pairs to align them using an LCS algorithm. By repeating this procedure recursively, two texts can be matched in an efficient way. Finally, all pairs are aligned along the pivot and a majority vote determines the final result.

\cite{liwicki2011combining} tackled the task of handwritten text recognition acquired from a whiteboard by combining several individual classifiers of diverse nature. They used two base recognizers which incorporated hidden Markov models and bidirectional LSTM networks and trained them on different feature sets. Moreover, two commercial recognition systems were added to the voting. The multiple classifier system reached an accuracy of 86.16\% on word level and therefore outperformed even the best individual system (81.26\%) significantly. 

\cite{al2015combination} trained neural LSTM networks on two OCR outputs aligned by weighted finite-state transducers based on edit rules in order to return a best voting. After training the network on a vast amount of data very similar to the test set, it was able to predict even characters which were not correctly recognized by either of the two engines. During tests on printings with German Fraktur and the UW-III data set the LSTM approach led to CERs of around 0.40\%, considerably outperforming the ISRI voting tool and the method presented in \cite{wemhoener2013creating} (between 1.26\% and 2.31\%). However, applying this method to historical spellings has a principal drawback as it relies on fixed input-output relationships. Since historical spelling patterns are much more variable than modern ones and the same word is often spelled and printed in more than one form even in the same document, it is not possible or at least may not be desired to map each OCR token to a single `correct' token.

In \cite{reul2017voting} we implemented a cross-fold training procedure with subsequent confidence voting in order to reduce the CER on early printed books. This method shows considerable differences compared to the work presented above. Not only is it applicable to some of the earliest printed books, but it also works with only a single open source OCR engine. Furthermore, it can be easily adapted to practically any given book using even a small amount of GT without the need for excessive data to train on (60 to 150 lines of GT corresponding to just a few pages will suffice for most cases).

By dividing the GT in $N$ different folds and aligning them in a certain way we were able to train $N$ strong but also diverse models. Then, these models acted as voters both in the default sequence voting (see ISRI tools above) and a newly created confidence voting scheme which also takes the intrinsic confidence information of the top-n (not just top-1) predictions of OCRopus into consideration. Experiments on seven books printed between 1476 and 1675 led to the following observations:

\begin{enumerate}
    \item For all experiments the cross fold training and voting approach led to significantly lower CERs compared to performing only a single training. Gains between 19\% and 53\% were reported for several books and different amount of lines of GT.
    \item OCR texts with a lower CER benefitted even more than more erroneous results.
    \item The amount of available GT did not show a notable influence on the improvements achievable by confidence voting. Yet, a very high number of lines leads to a drop in voting gains for most books. This has to be expected for models that get closer to perfection as most of the remaining errors are unavoidable ones such as characters with defects or untrained glyphs missing in the training set.
    \item Increasing the number of folds can bring down the CER even further, especially when training on a large set of lines. However, considering the range of available GT lines and the required computational effort, five folds appeared to be a sensible default choice until further testing regarding parameter optimization has been performed.
    \item The confidence voting always outperformed the standard sequence voting approach by reducing the amount of errors by another 5\% to 10\%.
\end{enumerate}

\subsection{Transfer Learning and OCR Pretraining}
\label{sec:RW_pt}
While to the best of our knowledge there is no suitable related work regarding transfer learning in the field of OCR, it was applied successfully to a variety of other tasks (e.g. \cite{yosinski2014transferable} for labeling arbitrary images and 
\cite{puppe2017} for leaf classification) by deploying deep convolutional neural networks after performing a pretraining on data from a different but somewhat similar recognition task.

Obviously, these examples of transfer learning used far deeper networks than OCRopus with only a single hidden layer, resulting in a dramatically increased number of parameters and consequently more opportunities to learn and remember useful low-level features. Nonetheless, since scripts in general should show a high degree of similarity we still expected a noteworthy impact of pretraining and studied the effect of building from an already available mixed model instead of starting training from scratch \citep[see][]{reul2017transfer}. As starting points we used the models for modern English, German Fraktur from the 19th century, and the Latin Antiqua model described above. From our experiments we were able to derive the following conclusions:

\begin{enumerate}
    \item Building from a pretrained model significantly reduced the obtainable CER compared to starting the training from scratch.
    \item Improvement rates decrease with an increasing amount of GT lines available for training. While models trained on only 60 lines of GT gained over 40\% on average over starting from scratch, this number went down to around 15\% for 250 lines.
    \item The incorporation of a whitelist for standard letters and digits which cannot be deleted from the codec even if they do not occur in the training GT showed an additional average gain of 5\%.
    \item Even the mixed models for modern English and 19th century Fraktur which were completely unrelated to the individual books in terms of printing type and age of the training material led to significant improvements compared to training from scratch.
\end{enumerate}

\subsection{Active Learning}
\cite{settles2012active} gives a very comprehensive overview over the literature dealing with Active Learning (AL). Apart from introducing different usage scenarios and discussing theoretical and empirical evidence for the application of AL techniques they define a typical AL scenario as follows: A learner starts out with access to a (possibly very small) pool of labeled examples to learn from. In order to improve performance it is possible to send queries consisting of one or several carefully selected unlabeled examples to a so-called oracle (teacher/human annotator) who then returns a label for each example in the query. Afterwards, the learner can utilize the obtained additional data. Obviously, the progress of the learner heavily depends on the examples selected to be labeled. Furthermore, the goal is to learn as much as possible from as few as possible queried examples, keeping the oracle's effort to a minimum.

One of the most successful query techniques is called \textit{query by committee} and was introduced by \cite{seung1992query}. The basic idea is that a committee of learners/models/voters is trained on the current labeled set. Each member of the committee is allowed to cast a vote on a set of query candidates (unlabeled examples). The assumption is that the candidate the voters disagree most on is also the one which offers the biggest information gain when being added to the training set. This is called the \textit{principle of maximal disagreement}.

Among others, the effect of this approach was demonstrated by \cite{krogh1995neural} who trained five neural networks to approximate the square-wave-function. They performed 2x40 independent test runs starting from a single example and using passive and active learning. While the next example was chosen randomly during the passive tests the networks always got handed the example with the largest ambiguity among the five voters out of 800 random ones. Evaluation showed that AL led to a significantly better generalization error and that the individual additional training examples on average contributed much more to the training process when chosen according to the principle of maximal disagreement.

As for OCR, \cite{springmann2016automatic} performed some initial experiments on selecting additional training lines in an active way. After recognizing lines with a mixed model they tested several strategies according to which they chose lines for further transcription. The best result was obtained when using a mixture of randomly selected lines combined with lines with low confidence values. It is worth mentioning that after transcribing these lines they started their training from scratch since the pretraining approach introduced above had not been developed, yet.

\section{Materials and Methods}
\label{section:matmet}
In this chapter we first introduce the early printed books and mixed models we used for our experiments. Then our previous approaches for separate voting and pretraining are briefly described on a technical level. Finally, we show how the principle of maximal disagreement can be utilized in order to choose additional training lines in an informed way within an iterative AL approach.

\subsection{Books}
The experiments were performed on six early printed books (see Table 1). The AL experiments were carried out on the three books above the horizontal line. We focused on these books as they provided a large amount of GT which is needed to perform the procedure. In a real world application scenario it would be sensible to choose the additional training lines by recognizing all lines without GT and choose the worst ones. Therefore, as many lines as possible are required to be able to evaluate this scenario.

\begin{table}[tb]
\centering
\caption{Books used during the experiments as well as the amount of GT lines set aside for Training, Evaluation, and Active Learning.}
\label{tab:books}
\begin{tabular}{ccccc}
\hline\noalign{\smallskip}
\textbf{ID/Year} &	\textbf{Language} &	\textbf{Training} & \textbf{Evaluation} & \textbf{Active Learning}\\
\noalign{\smallskip}\hline\noalign{\smallskip}
1476 & German & 1,000 & 1,000 & 750\\
1488 & German & 1,000 & 1,000 & 1,928\\
1505 & Latin &	1,000 &	1,000 & 1,039\\
\hline
1495 & German &	1,000 &	1,114 & -\\
1500 & Dutch &	1,000 &	1,250 & -\\
1572 &	Latin &	1,000 &	1,098 & -\\
\hline
\end{tabular}
\end{table}

To avoid unwanted side effects resulting from different types or varying line length only lines from running text were used and headings, marginalia, page numbers, etc. were excluded. 1505 represents an exception to that rule as we chose the extensive commentary lines instead, as they presented a bigger challenge due to very small inter character distances and a higher degree of degradation. Figure 1 shows some example lines.

\begin{figure}[tb]
\centering
\includegraphics[width = 0.65\linewidth]{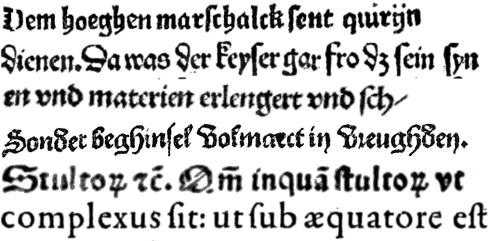}
\caption{Different example lines from the six books used for evaluation.
From top to bottom: excerpts from books 1476, 1488, 1495, 1500, 1505, and 1572.}
\label{figure:lines}
\end{figure}

The books published in 1495, 1500, and 1505 are editions of the Ship of Fools \textit{Narrenschiff} and were digitized as part of an effort to support the Narragonien project at the University of Würzburg\footnote{\url{http://kallimachos.de/kallimachos/index.php/Narragonien}}. Despite their similar content these books differ considerably from an OCR point of view since they have been printed in different print shops using different typefaces and languages (Latin, German, and Dutch, see Figure~\ref{figure:lines}). Ground truth for the 1488 book was gathered during a case study of highly automated layout analysis \cite[see][]{reul2017dhl}. 1476 is part of the Early New High German Reference Corpus\footnote{\url{http://www.ruhr-uni-bochum.de/wegera/ref/index.htm}} and 1572 was digitized in order to be added to the AL-Corpus\footnote{\url{http://arabic-latin-corpus.philosophie.uni-wuerzburg.de}} of Arabic-Latin translations.

\subsection{Mixed Models}
During our experiments we made use of three mixed models. Our first model LH (abbreviated for Latin Historical) was trained on all twelve historical books introduced in \cite{springmann2016automatic}. The books are printed in Latin and with Antiqua types. The training was performed on 8,684 lines and was stopped after 109,000 iterations. We evaluated all resulting models on 2,432 previously unseen test lines in order to determine the best model which occurred after 98,000 training steps achieving a CER of 2.92\% .

Additionally, we used the freely available OCRopus standard models for English (ENG)\footnote{\url{http://www.tmbdev.net/en-default.pyrnn.gz}} and German Fraktur (FRK)\footnote{\url{http://tmbdev.net/ocropy/fraktur.pyrnn.gz}} introduced in \cite{breuel} and described above. 

\subsection{Cross Fold Training and Confidence Voting}
In the absence of valid alternatives to OCRopus we introduced variations in the training data in order to obtain highly performant individual yet diverse voters. This was done by applying a technique called cross fold training: The available GT is divided into $N$ folds with $N$ being the number of models which will later participate in the vote. Then, $N$ training processes take place by using one fold for testing, i.e. choosing the best model, and the rest for training. While the training data shows a significant overlap for each training the folds used for testing are distinct.

After determining the best model of each training process each one of the resulting best models recognizes the same set of previously unknown lines. As an output not only the OCR text is stored but also so-called \textit{extended lloc} (\textbf{L}STM \textbf{l}ocation \textbf{o}f \textbf{c}haracters) files which store the probability of the actually recognized character as well as the probabilities of its alternatives.

During the confidence voting process the different result texts are first aligned by applying the ISRI sync tool which identifies the positions of OCR differences at as well as the output of each voter at this position. The confidence voting is performed by identifying the corresponding extended llocs and adding up the confidence values for each of the character alternatives. The character with the highest confidence sum gets voted into the final output \citep{reul2017voting}.

\subsection{Pretraining utilizing Transfer Learning}
The character set of pretrained mixed models often comprises more or less characters than the data the new model is supposed to be trained on. In order to assure full compatibility we had to make some enhancements on code level regarding the OCRopus training process which are available at GitHub\footnote{\url{https://github.com/ChWick/ocropy/tree/codec_resize}}. These changes allow to comfortably extend or reduce the codec depending on the available training GT and the characters it contains. When starting the training process the character sets of the existing model and the GT are matched. For each character which occurs in the GT but is not part of the model an additional row is attached to the weight matrix and the corresponding weights are randomly initialized. A character which is known to the model but is not part of the GT leads to the deletion of the corresponding row in the matrix. In order to avoid blind spots especially when dealing with small amounts of GT and important but less frequent characters like numbers or capital letters it is possible to define a so-called \textit{whitelist}. Characters on the whitelist will not get deleted from the matrix no matter if they occur within the GT or not \citep{reul2017transfer}.

\subsection{Active Learning}
\label{sec:MM_AL}
As explained above the basic idea behind Active Learning is to allow the learners, i.e. the different voters (see \ref{sec:MM_AL}), to decide which training examples they benefit the most from instead of selecting additional lines randomly. Since in a real world application scenario there usually is no GT available for potential new training lines, we cannot just use the ones for which our current models give the worst results, i.e. the ones with the highest CER. However, we can still identify the lines we expect to be most suitable for further training by following the principle of maximal disagreement.

We utilize the normalized Levenshtein distance ratio (LDR) as our measure for the ambiguity between the OCR results obtained from the available models: each output is compared to the outputs of the other voters one by one, the individual edit distance is computed, divided the length of the text line and added up for each individual line. Then, the text lines are sorted in descending order according to the sum of their LDRs. In a real world application scenario the lines are handed to a human annotator who then produces GT by transcribing them or by correcting one of the individual OCR results or the output of the confidence voting, if available. While the idea of the whole process is to give the committee the lines it requested, it is still up to the human annotator to decide whether a line is suitable for further training or not. For example, a line might have been badly recognized due to a severe segmentation error or due to an unusually high degree of degradation making recognition pretty much impossible. These lines cannot be expected to make a noteworthy contribution to the training process and are therefore discarded.

\section{Experiments}
\label{section:exp}
In order to evaluate the effectiveness of the methods described above we performed two main experiments. First, the voting and pretraining approaches were combined\footnote{The corresponding code is available at GitHub: \url{https://github.com/chreul/mptv}} by performing the voting procedure with models which were not trained from scratch but started from one or several pretrained mixed models. Second, the voters resulting from the first experiment served as a committee during an AL approach following the principle of maximal disagreement.

Since the train/test/evaluation distribution of the GT lines has changed, the results can differ from the ones obtained from earlier experiments. Based on the previous results reported in sections \ref{sec:RW_vote} and \ref{sec:RW_pt} we chose to implement the following guidelines for all of the upcoming experiments.

\begin{enumerate}
    \item The number of folds during cross-fold training and consequently the number of voters is set to 5.
    \item OCR results are always combined by enforcing the confidence voting approach.
    \item Whenever pretraining is used a generic minimal whitelist consisting of the letters a-z, A-Z and the numbers 0-9 is added to the codec.
    \item Each model training is carried out until no further improvement is expected (e.g. 30,000 iterations for 1,000 lines of training GT).
\end{enumerate}

\subsection{Combining Pretraining and Voting}
Obviously, the combination of voting and pretraining seems attractive and should be evaluated. The amount of lines used for training was varied in six steps from 60 to 1,000. Each set of lines was divided in five folds and the allocation was kept fixed for all experiments. We used two different approaches for pretraining. First, we trained the five voters by always building from the LH model since it yielded the best results during previous experiments. Second, we varied the models used for pretraining. We kept voters 1 and 2 from the first setup (trained from LH). For voters 3 and 4 FRK was utilized as a starting point since it was trained on German Fraktur fonts which are somewhat similar to the broken script of the books at hand. Only one voter (5) was built from ENG as it was the least similar one out of the available mixed models regarding both age and printing type of the training data. This setup was slightly adapted for book 1572 as it was printed using Antiqua types. Therefore, in this case one of the two FRK folds was pretrained with ENG instead.

\begin{figure}[t]
\centering
\includegraphics[width=\linewidth]{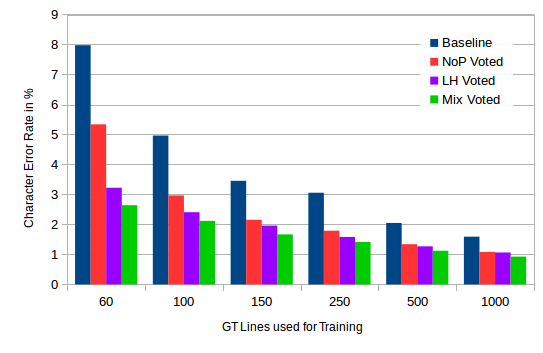}
\caption{Comparison of the CERs (averaged over all books for each set of lines) of four different approaches: \textit{Baseline} (no pretraining, no voting), \textit{NoP} (no pretraining, voted), \textit{LH} (all five folds trained from the LH model, voted), and \textit{Mix} (mixed pretraining, voted).}
\label{fig:avgRes}
\end{figure}

\begin{figure}[t]
\centering
\includegraphics[width=\linewidth]{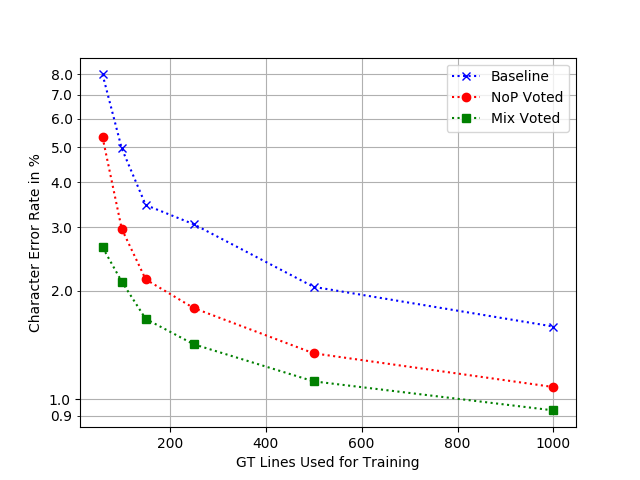}
\caption{Influence of the number of GT training lines compared for the approaches \textit{Baseline} (no pretraining, no voting), \textit{NoP Voted} (no pretraining, voted), and \textit{Mix Voted} (pretrained with different mixed models, voted) on a logarithmic scale for CER.}
\label{fig:lineBehaviour}
\end{figure}

The idea was to still train strong individual models while increasing diversity among them, hoping for a positive effect on the final voting output. The results are summed up in Table \ref{tab:ptVote}. For reasons of clarity, detailed numbers are only provided for three books, i.e. the ones which will be used for further experiments. The general behaviour averaged over all books can be seen in Figure \ref{fig:avgRes} and an overview over the progress made by adding more training lines is presented in Figure \ref{fig:lineBehaviour}.

\begin{table}[h!]
\centering
\caption{CER in \% of combining pretraining (\textit{Single Folds}) and voting (\textit{Voting Result}). \textit{Single Folds} contain the baseline without pretraining  (\textit{Base}), pretraining with LH model (\textit{LH}), and with a mixture of models (LH, LH, FRK, FRK/ENG, ENG) (\textit{Mix}). \textit{Voting Result} shows the results of different voters based on no pertraining  (\textit{NoP}), pretraining with LH model (\textit{LH}), and with mixed model (\textit{Mix}). The \textit{Improvement} columns show the voting gains of LH over NoP (\textit{NL}), Mix over NoP (\textit{NM}), Mix over LH (\textit{LM}), and Mix over the base (\textit{BM}). The underlined CERs represent the starting points for the upcoming AL experiment.}
\label{tab:ptVote}
\begin{tabular}{c|ccc|ccc|cccc}
\hline\noalign{\smallskip}
\textbf{\underline{1476}}  & \multicolumn{3}{c|}{\textbf{Single Folds}} & \multicolumn{3}{c|}{\textbf{Voting Result}} & \multicolumn{4}{c}{\textbf{Improvement}}                                                 \\
\textbf{Lines} & \textbf{Base}   & \textbf{LH}   & \textbf{Mix}  & \textbf{NoP}  & \textbf{LH} & \textbf{Mix} & \textbf{NL} & \textbf{NM} & \textbf{LM} & \textbf{BM} \\
\hline
\textbf{60}    & 8.12           & 5.58          & 4.96          & 4.72          & 4.10        & 2.79         & 13\%                        & 41\%                         & 32\%        & 66\%                \\
\textbf{100}   & 6.82           & 3.99          & 3.92          & 3.49          & 2.67        & \underline{2.23}         & 23\%                        & 36\%                         & 16\%        & 67\%                \\
\textbf{150}   & 4.10           & 3.15          & 3.03          & 2.47          & 2.14        & 1.66         & 13\%                        & 33\%                         & 22\%       & 60\%                 \\
\textbf{250}   & 3.24           & 2.40          & 2.37          & 1.70          & 1.63        & \underline{1.47}         & 4\%                         & 14\%                         & 10\%        & 55\%                \\
\textbf{500}   & 2.11           & 1.73          & 1.75          & 1.17          & 1.13        & 1.03         & 3\%                         & 12\%                         & 9\%     & 51\%                    \\
\textbf{1000}  & 1.55           & 1.30          & 1.22          & 0.97          & 0.88        & 0.75         & 9\%                         & 23\%                         & 15\%    & 52\% \\
\hline
\hline
\textbf{\underline{1488}} & \multicolumn{3}{c|}{\textbf{Single Folds}} & \multicolumn{3}{c|}{\textbf{Voting Result}} & \multicolumn{4}{c}{\textbf{Improvement}}             \\
\textbf{Lines}      & \textbf{Base}   & \textbf{LH}  & \textbf{Mix}  & \textbf{NoP}  & \textbf{LH} & \textbf{Mix} & \textbf{NL} & \textbf{NM} & \textbf{LM}  & \textbf{BM} \\
\hline
\textbf{60}         & 7.28           & 3.97         & 4.28          & 4.38          & 3.05        & 2.50         & 30\%            & 43\%             & 18\%    & 66\%        \\
\textbf{100}        & 4.19           & 2.85         & 3.20          & 2.73          & 2.06        & \underline{1.84}         & 25\%            & 33\%             & 11\%    & 56\%        \\
\textbf{150}        & 2.96           & 2.26         & 2.33          & 1.81          & 1.51        & 1.24         & 17\%            & 31\%             & 18\%    & 58\%        \\
\textbf{250}        & 2.59           & 1.82         & 1.89          & 1.29          & 1.21        & \underline{1.07}         & 6\%             & 17\%             & 12\%    &59\%        \\
\textbf{500}        & 1.50           & 1.40         & 1.38          & 0.91          & 0.95        & 0.79         & -4\%           & 13\%             & 17\%     & 47\%       \\
\textbf{1000}       & 1.17           & 1.06         & 1.13          & 0.71          & 0.72        & 0.61         & -1\%            & 14\%             & 15\%    & 48\% \\
\hline
\hline
\textbf{\underline{1505}} & \multicolumn{3}{c|}{\textbf{Single Folds}} & \multicolumn{3}{c|}{\textbf{Voting Result}} & \multicolumn{4}{c}{\textbf{Improvement}}             \\
\textbf{Lines}      & \textbf{Base}   & \textbf{LH}  & \textbf{Mix}  & \textbf{NoP}  & \textbf{LH} & \textbf{Mix} & \textbf{NL} & \textbf{NM} & \textbf{LM} & \textbf{BM} \\
\hline
\textbf{60}         & 6.54           & 5.00         & 5.27          & 4.58          & 3.70        & 3.45         & 19\%            & 25\%             & 7\%     & 47\%       \\
\textbf{100}        & 4.54           & 3.96         & 4.15          & 3.16          & 2.82        & \underline{2.68}         & 11\%            & 15\%             & 5\%     & 41\%        \\
\textbf{150}        & 3.54           & 3.16         & 3.16          & 2.34          & 2.27        & 2.02         & 3\%             & 14\%             & 11\%    & 43\%        \\
\textbf{250}        & 2.85           & 2.66         & 2.18          & 1.98          & 1.77        & \underline{1.60}         & 11\%            & 19\%             & 10\%    & 44\%         \\
\textbf{500}        & 2.24           & 2.18         & 2.11          & 1.59          & 1.60        & 1.43         & -1\%            & 10\%             & 11\%    &36\%        \\
\textbf{1000}       & 1.84           & 1.85         & 1.82          & 1.35          & 1.40        & 1.26         & -4\%            & 7\%              & 10\%    & 32\% \\
\hline
\hline
\hline
\textbf{\underline{All}} & \multicolumn{3}{c|}{\textbf{Single Folds}} & \multicolumn{3}{c|}{\textbf{Voting Result}} & \multicolumn{4}{c}{\textbf{Improvement}}             \\
\textbf{Lines}      & \textbf{Base}   & \textbf{LH}  & \textbf{Mix}  & \textbf{NoP}  & \textbf{LH} & \textbf{Mix} & \textbf{NL} & \textbf{NM} & \textbf{LM} & \textbf{BM} \\
\hline
\textbf{60}         & 7.98           & 4.42         & 4.49          & 5.34          & 3.22        & 2.64         & 40\%            & 51\%             & 18\% & 67\%            \\
\textbf{100}        & 4.97           & 3.38         & 3.49          & 2.97          & 2.41        & 2.12         & 19\%            & 29\%             & 12\%    & 57\%         \\
\textbf{150}        & 3.46           & 2.78         & 2.87          & 2.15          & 1.96        & 1.67         & 9\%             & 23\%             & 15\%    & 52\%        \\
\textbf{250}        & 3.06           & 2.28         & 2.20          & 1.79          & 1.59        & 1.42         & 11\%            & 21\%             & 11\%    & 54\%         \\
\textbf{500}        & 2.05           & 1.86         & 1.77          & 1.34          & 1.27        & 1.12         & 5\%            & 16\%             & 12\%     & 45\%       \\
\textbf{1000}       & 1.59           & 1.46         & 1.42          & 1.08          & 1.07        & 0.93         & 2\%            & 14\%              & 13\%    & 42\% \\
\hline
\textbf{Avg.}       & 3.85           & 2.70         & 2.69          & 2.45          & 1.92        & 1.65         & 14\%            & 26\%              & 13\%  & 53\%\\
\hline
\end{tabular}
\end{table}

In the majority of cases the combination of pretraining and voting considerably outperforms both the default voting approach showing gains of 14\% (LH) and 26\% (Mix) as well as the default pretraining approach showing gains of 29\% (LH) and 39\% (Mix) when averaging over all six books and lines. As expected, the best improvements can be achieved when using a small number of GT lines, resulting in gains ranging from 40\% (LH) and 51\% (Mix) for 60 lines to 2\% (LH) and 14\% (Mix) for 1,000 lines.

Overall, the average CER on individual folds pretrained with LH (2.70\%) is effectively identical to the one achieved by building from a variety of mixed models (2.69\%). However, in case of applying the voting procedure over all five folds, the Mix approach yields considerably better results than just using LH a a pretrained model, leading to an additional reduction of recognition errors by over 13\% without an apparent correlation regarding the amount of training GT.

Comparing the best method (Mix + Voting) with the baseline, i.e. the default OCRopus approach (training a single model without any pretraining or voting), shows the superiority of the proposed approach yet more clearly. Even the error rates of strong individual models trained on a 1,000 lines of GT are reduced by more than 40\% on average. In general, a substantial amount of GT (>250 lines) is required in the standard OCRopus training (see `baseline' in Figure \ref{fig:avgRes}) to match the result achieved by a mere 60 lines when incorporating mixed pretraining and voting, indicating a GT saving factor of 4 or more. A similar factor can be derived when considering the average baseline CER for 1,000 lines of GT.

\subsection{Incorporating Active Learning}
To select additional training lines we utilized the models (voters) obtained from the previous experiment. Each model recognized the GT lines set aside for AL and the best candidates were determined by choosing the ones with the highest LDR sum as explained in section \ref{sec:MM_AL}. Next, the candidates underwent a quick visual inspection in order to sort out lines where a positive impact on the training was considered highly unlikely due to very rare segmentation errors or extreme degradations. Obviously, this adds a bit of subjectivity to the task but we expected the decision whether to keep or drop a line to be trivial in most cases. However, in our experiments we never skipped a line proposed by the AL approach despite coming across several borderline cases which will be discussed below. 

We performed two experiments starting with different amounts of base lines (100/250) which we kept from the previous experiment. For the passive learning approach we added an additional 50\% (50/125) of randomly selected lines. This was performed five times and the results were averaged. As for AL we chose the lines by following the principle of maximal disagreement incorporating the LDR. Since lines have different amounts of letters we paid attention to select only as many lines as necessary to match the average amount of characters in the passive learning approach.

After selecting the lines the base fold setup was kept and we distributed the additional GT evenly over the five folds to ensure an effect on the training itself but also on the selection of the best model. Afterwards, the training was started from scratch/from the default mixed models while the voters were discarded.

Since the previous experiment showed the superiority of the mixed pretraining approach we decided to omit the pretraining using LH during the upcoming experiments. Table \ref{tab:al} shows the results.

\begin{table}[t]
\centering
\caption{Results of comparing active to passive learning. \textit{Base Lines} is the number of lines used to train the voters of the previous iteration. The lines added (\textit{Add. Lines}) randomly (\textit{RDM}) or by maximizing the LDR (\textit{AL}) correspond to 50\% of the base lines. Compared are the resulting error rates (\textit{CER}) after performing a confidence vote and (in case of RDM) an averaging calculation. Finally, (\textit{Avg. Gain}) shows the average improvement of the voters trained by AL.}
\label{tab:al}
\begin{tabular}{c|cc|c|ccc}
\hline\noalign{\smallskip}
\multirow{2}{*}{\textbf{Book}} & \multicolumn{2}{c|}{\textbf{Base}} & \multirow{2}{*}{\textbf{\begin{tabular}[c|]{@{}c@{}}Add. Lines\\ Rdm/AL\end{tabular}}} & \multicolumn{2}{c}{\textbf{CER}} & \multirow{2}{*}{\textbf{\begin{tabular}[c]{@{}c@{}}Average\\ Gain\end{tabular}}} \\
                               & \textbf{Lines}   & \textbf{CER}   &                                                                                          & \textbf{Rdm}    & \textbf{AL}    &                                                                                  \\
                               \hline
\multirow{2}{*}{\textbf{1476}} & 100              & 2.23           & 50                                                                                       & 1.80            & 1.31           & 26\%                                                                             \\
                               & 250              & 1.47           & 125                                                                                      & 1.18            & 0.90           & 24\%                                                                             \\
\hline
\multirow{2}{*}{\textbf{1488}} & 100              & 1.84           & 50                                                                                       & 1.54            & 1.05           & 32\%                                                                             \\
                               & 250              & 1.07           & 125                                                                                      & 0.86            & 0.65           & 24\%                                                                             \\
                               \hline
\multirow{2}{*}{\textbf{1505}} & 100              & 2.68           & 50                                                                                       & 2.17            & 2.21           & -3\%                                                                              \\
                               & 250              & 1.81           & 125                                                                                      & 1.60            & 1.57           & 0\%       \\
                            \hline
\end{tabular}
\end{table}

Incorporating AL leads to lower CERs for four out of six tested scenarios. While important improvements with an average gain of almost 27\% can be reported for 1476 and 1488, 1505 does not improve at all (see the discussion in the next section).

Moreover, it is worth mentioning that no clear influence of the amount of the GT lines available for training can be derived from these results. Even when starting from an already quite comprehensive GT pool of 250 lines AL yielded an average gain of 16\% compared to randomly chosen lines.

\begin{figure}[t]
\centering
\includegraphics[width=\linewidth]{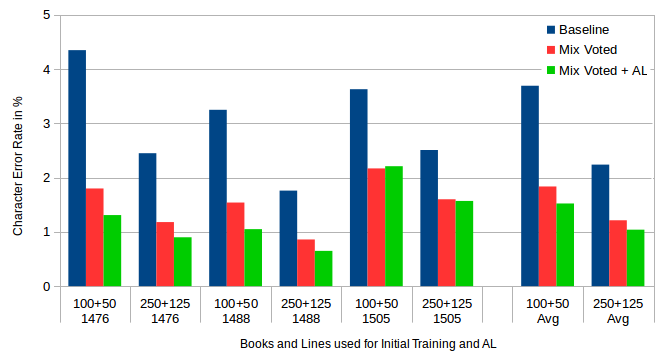}
\caption{Results of the AL experiments for three books at two different sets of lines comparing the \textit{Baseline} (no pretraining, no voting), \textit{Mix Voted} (pretrained with different mixed models, voted) and \textit{Mix Voted + AL} (Mix with additional lines chosen by AL).}
\label{fig:finalComp}
\end{figure}

Finally, Figure \ref{fig:finalComp} sums up the results by comparing the baseline to the best pretraining (Mix) approach combined with confidence voting with and without AL.

\section{Discussion}
\label{section:disc}
The experiments show that the combination of pretrained models and confidence voting is an effective way to further improve the achievable CER on early printed books. While the obtainable gain is highest when the number of available GT lines is small, a substantial reduction of OCR errors can still be expected even when training with several hundreds of lines. 

An interesting result of our experiments is that the variability of the voters has a clear influence on the voting result and can even outweigh a superior individual quality of the single voters. This explains why using a variety of models for pretraining considerably outperformed $N$-fold training from the LH model even though it represented the best fitting one of the available mixed models. Since training from an available model skips the random initialization of values in the weight matrix for pre-existing characters an important chance of introducing diversity to the training process is skipped, resulting in quite similar models even when trained on different but still heavily overlapping folds of training GT.

Following the proposed approach of combining a mixed pretraining with confidence voting allows for a substantially more efficient use of the available GT. On average we were able to achieve the same results as the standard OCRopus approach requiring less than one fourth of GT lines to do so. Moreover, a tiny amount of 60 lines was enough to reach an average CAR of close to 97.5\%. Only two out of six books showed a CER of over than 3\% but comfortably surpassed this value when adding another 40 lines of GT, raising the average to almost 98\% character accuracy, which is already considered good enough for many areas of application.

\begin{figure}[t]
\centering
\includegraphics[width = 0.8\linewidth]{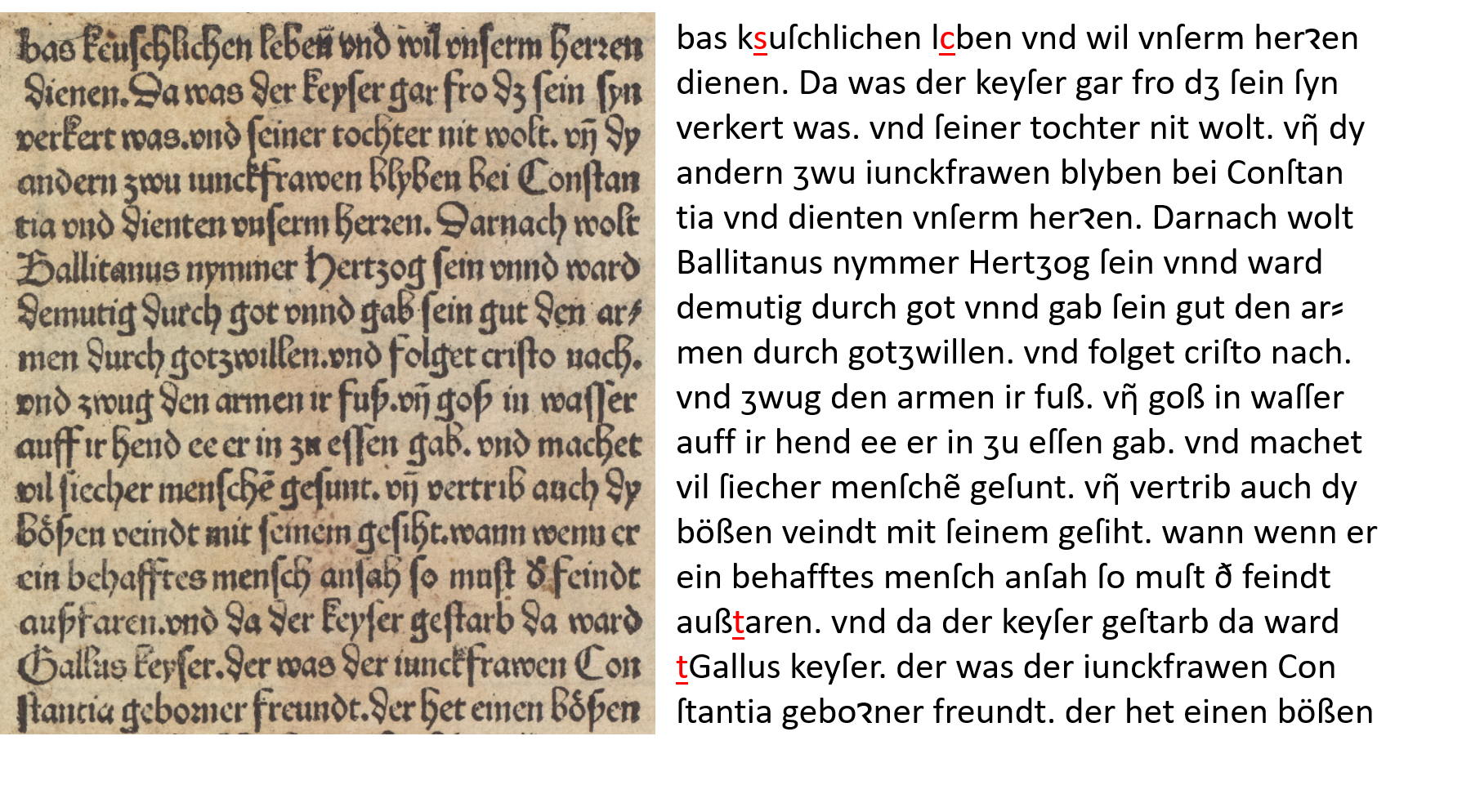}
\caption{Snippet of a scanned page for 1488 and its best OCR output by combining mixed pretraining and voting. The resulting CER of 0.60\% was achieved by training on 1,000 GT lines. The remaining four recognition errors are marked in red and underlined.}
\label{figure:DHL}
\end{figure}

Notwithstanding, there are opportunities for optimizing the achieved results even further, e.g. in applications where the goal is to manually check and correct an entire book in order to obtain a CER of close to 0\%. The experiments showed that our method also significantly outperforms the standard approach when training on a very comprehensive GT pool of 1,000 lines, resulting in an average CER of less than 1\%. Exemplary, Figure~\ref{figure:DHL} gives an impression of input (scanned page image) and output (best possible OCR result) for the 1488 printing with a CER of 0.60\% reached by training on 1,000 lines combining pretraining using a variety of mixed models and confidence voting.

Obviously, even when it is intended to transcribe an entire book the goal still should be to minimize the CER by investing the least possible manual correction effort. Therefore, an iterative training approach makes sense and a sensible selection of further training lines is important. Our experiments on AL showed that choosing additional lines in an informed way can be utilized to obtain an even more efficient way to use the available GT. Despite one of the books not responding at all to the proposed method due to heavily and inconsistently degraded glyphs the three evaluated books still showed an improvement of 17\% compared to the mixed voting approach when selecting additional training lines randomly for transcription. It is worth mentioning that during our experiments the amount of lines presented to the committee of voters was considerably smaller than in a real world application scenario where it might be sensible to take all yet unknown lines into consideration in order to choose the most promising ones in terms of information gain. This might have a positive effect on the achievable results of the AL approach.

\begin{figure}[t]
\centering
\includegraphics[width=\linewidth]{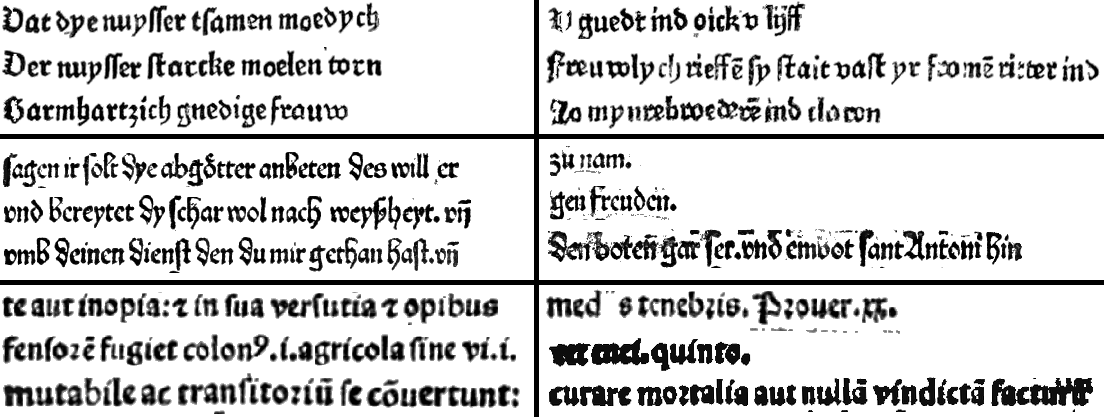}
\caption{Example lines of the three books 1476 (top), 1488 (middle), and 1505 (bottom) which were presented to the committee. The left column shows perfectly recognized lines (average LDR of 0.0). On the right some of the most erroneous lines are presented (average LDR of 0.45 (1476), 0.72 (1488), and 0.64 (1505)).}
\label{fig:AL_lines}
\end{figure}

A likely explanation for the lack of improvements of the 1505 book by AL is the degree and type of degradations of the lines queried for training by the voting committee which is illustrated by some example lines shown in Figure \ref{fig:AL_lines}. The lines on the left are examples the voters fully agreed on and, as expected, got recognized correctly by the base models trained with 100 lines of GT. On the right some of the lines are shown where the committee disagreed the most on, i.e. the ones with highest LDR. For 1476 (top) the lines shown had a ratio of 0.45 and CER of 9.44\%. There are some signs of degradation, mostly moderately faded glyphs. The worst lines selected from 1488 (middle, 0.72 LDR, 33.68\% CER) mostly suffered from noise while the glyphs of 1505 (bottom, 0.64 LDR, 30.77\% CER) frequently show severe deformations. This might be a sensible explanation why AL works very well for 1476 and 1488 but not at all for 1505. Despite the fading and the noise the glyphs of 1476 and 1488 look much more regular than the deformed ones of 1505, at least to the human eye. Therefore, the models trained for 1476 and 1488 using AL learned to see through the effects of fading and noise and earned additional robustness, resulting in a considerable gain in CER. In the case of 1505 the AL models were fed many lines showing severe but very irregular degradations which may have led to an increased robustness but probably did not improve the recognition capability of regular lines as much as the passive learning lines did.

\section{Conclusion and Future Work}
\label{section:con}
In this paper we proposed a combination of pretraining, confidence voting, and AL in order to significantly improve the achievable CER on early printed books. The methods were shown to be very effective for different amounts of GT lines typically available from several hours' transcription work of early printings.

In the future we aim to utilize the positive effect of more diverse voters even further. In general, a higher degree of diversity can be achieved in two ways:

\begin{enumerate}
\item By the inclusion of more mixed models for pretraining. Obviously, since there are only a few good mixed models freely available to date, there is a great need and sharing is key. To lead by example we made part of our GT data and models available online\footnote{\url{https://github.com/chreul/OCR_Testdata_EarlyPrintedBooks}} and more mixed models are provided in the companion paper by Springmann et al. on \textit{Ground Truth for OCR on historical documents} (this issue).

\item By varying the network structure used for training representing a viable leverage point to increase diversity even further. First steps in this direction have recently been made by \cite{breuel17hybrid} and Wick et al. (\textit{Improving OCR Accuracy on Early Printed Books using Deep Convolutional Networks}; submitted to this issue.).
\end{enumerate}

Obviously, the combination of both intended improvements also represents a very promising approach.

To optimize the achievable results, extensive experiments regarding parameter optimization are required. This includes the number of folds/voters, the kind of network they have been trained on, and the method of combination of mixed models used for pretraining, as well as the number of lines the training is performed on.

As for AL an important additional approach is to not only utilize it in order to choose new training lines before the actual training process but also to get involved during the training itself. The standard OCRopus approach is to randomly select lines and feed them to the network. A more efficient method might be to decrease the chance to get picked for lines which already got perfectly recognized and consequently increase it for lines which still cause the current model a lot of problems.

Concerning the maximal disagreement approach for line selection, it would be interesting to experiment with other ways to determine those lines that offer a maximal information gain. Utilizing the intrinsic OCRopus confidence values that we already use during our voting approach comes to mind but also metrics like the Kullback-Leibler-Divergence.

Finally, despite our focus on early printed books the proposed methods are applicable to newer works as well. Especially 19th century Fraktur presents an interesting area of application. Since the typically used Fraktur typesets are more regular than those of the books used during our experiments the goal is to produce a mixed model with excellent predictive power and to avoid book specific training at all.

Nevertheless, the present paper is just a first step to the larger goal of creating an effective, open-source, computer-assisted OCR workflow that is automized to the largest extent possible. In the context of the OCR-D project this also encompasses methods of page segmentation \citep{reul2017larex} in the preprocessing phase as well as postcorrection \citep{fink2017profiling} in the postprocessing stage which are just as important for the quality of the end result as the recognition process itself. To enable further progress in the direction of better recognition results and better polyfont models, we made an extensive set of historical GT data of almost 300,000 lines available (see the article of Springmann, Reul, Dipper, Baiter in this issue).

\nocite{*}
\bibliographystyle{apa}
{\small \bibliography{references}}

\end{document}